\documentclass[journal]{IEEEtran}
\usepackage{amssymb}
\usepackage{amsmath,amsfonts}
\usepackage{algorithmic}
\usepackage{array}
\usepackage[caption=false,font=normalsize,labelfont=sf,textfont=sf]{subfig}
\usepackage{textcomp}
\usepackage{stfloats}
\usepackage{url}
\usepackage{verbatim}
\usepackage{xcolor}
\usepackage{graphicx}

\usepackage{algorithm}
\usepackage{booktabs}
\usepackage{multirow}
\usepackage{bbding}
\usepackage{pifont}
\usepackage{wasysym}
\usepackage[colorlinks=true, linkcolor=blue, citecolor=blue, urlcolor=blue]{hyperref}
\usepackage{newfloat}
\usepackage{cite}

\begin{document}

\title{Parameter Aware Mamba Model for Multi-task Dense Prediction}
\author{Xinzhuo Yu, Yunzhi Zhuge,~\IEEEmembership{Member,~IEEE}, Sitong Gong, Lu Zhang,  Pingping Zhang,~\IEEEmembership{Member,~IEEE}, and Huchuan Lu,~\IEEEmembership{Fellow,~IEEE}

\thanks{Xinzhuo Yu is with the School of Computer Science, Dalian University of Technology, Dalian 116081, China. (e-mail: yxz\_3132@mail.dlut.edu.cn)}
\thanks{Yunzhi Zhuge, Sitong Gong and Lu Zhang are with the School of Information and Communication Engineering, Dalian University of Technology, Dalian 116081, China. (e-mail: zgyz@dlut.edu.cn; stgong@mail.dlut.edu.cn; zhangluu@dlut.edu.cn)}
\thanks{Pingping Zhang is with the School of Future Technology and Artificial Intelligence, Dalian University of Technology, Dalian 116081, China. (email: zhpp@@dlut.edu.cn)}
\thanks{Huchuan Lu is with the School of Information and Communication Engineering, Dalian University of Technology, Dalian 116081, China, also with the School of Future Technology and Artificial Intelligence, Dalian University of Technology, Dalian 116081, China. (email: lhchuan@dlut.edu.cn)}
\thanks{Xinzhuo Yu and Yunzhi Zhuge contribute equally to this work.}
\thanks{Corresponding author: Huchuan Lu}}

\markboth{IEEE Transactions on Cybernetics}%
{Shell \MakeLowercase{\textit{et al.}}: A Sample Article Using IEEEtran.cls for IEEE Journals}

\maketitle

\begin{abstract}
Understanding the inter-relations and interactions between tasks is crucial for multi-task dense prediction. Existing methods predominantly utilize convolutional layers and attention mechanisms to explore task-level interactions. In this work, we introduce a novel decoder-based framework, Parameter Aware Mamba Model (PAMM), specifically designed for dense prediction in multi-task learning setting. 
Distinct from approaches that employ Transformers to model holistic task relationships, PAMM leverages the rich, scalable parameters of state space models to enhance task interconnectivity. It features dual state space parameter experts that integrate and set task-specific parameter priors, capturing the intrinsic properties of each task. This approach not only facilitates precise multi-task interactions but also allows for the global integration of task priors through the structured state space sequence model (S4). Furthermore, we employ the Multi-Directional Hilbert Scanning method to construct multi-angle feature sequences, thereby enhancing the sequence model's perceptual capabilities for 2D data. Extensive experiments on the NYUD-v2 and PASCAL-Context benchmarks demonstrate the effectiveness of our proposed method. Our code is available at \href{https://github.com/CQC-gogopro/PAMM}{https://github.com/CQC-gogopro/PAMM}. 
\end{abstract}

\begin{IEEEkeywords}
Dense Prediction, State Space Model, Multi-task Learning, Mixture of Experts.
\end{IEEEkeywords}

\section{Introduction}
\IEEEPARstart{I}{nspired by} cognitive science and psychology~\cite{posner1990attention}, which show that the human brain processes tasks by integrating various information sources, including sub-aspects of the task and prior knowledge~\cite{fuster2001prefrontal}, Multi-Task Learning (MTL) in computer vision adopts a similar integrative approach. By simultaneously handling multiple tasks, MTL allows information from one task to enrich and enhance another, leveraging shared representations to improve computational efficiency and model robustness.

As a machine learning paradigm, MTL concurrently performs multiple related yet distinct tasks within a single model~\cite{zhang2021survey}. This methodology is particularly advantageous in computer vision, where tasks often share intrinsic correlations at the feature level. MTL optimizes resource utilization through network sharing and incorporates task interaction modules that facilitate communication between tasks, thereby enhancing overall performance. This paper focuses on MTL for dense prediction, a fine-grained perception task aimed at pixel-level label prediction. Dense prediction is crucial for applications in complex environments such as autonomous driving and embodied intelligence, where accurate and detailed environmental understanding is essential.

The advancements in convolutional neural networks (CNNs) and Transformers have significantly enhanced the capabilities of multi-task learning (MTL). Initially, CNN-based methods such as NDDR-CNN~\cite{gao2019nddr} and Cross-Stitch~\cite{Misra_Shrivastava_Gupta_Hebert_2016} facilitated localized task feature sharing, which promoted effective task interactions. Following this, PAD-Net~\cite{xu2018pad} introduced a novel approach by supervising a set of auxiliary tasks to generate predictions, which were then used as multi-modal inputs for the final task. Similarly, MTI-Net~\cite{vandenhende2020mti} explicitly modeled task interactions across multiple scales, effectively refining task information from lower to higher levels.

Building on these CNN-based methods, Transformer-based approaches have been widely adopted in MTL due to their global interaction capabilities, demonstrating promising results~\cite{ye2022invpt,zhang2023demt,ye2024invpt++,xu2023deformable}. For example, InvPT~\cite{ye2022invpt} and InvPT++~\cite{ye2024invpt++} optimized multi-task information utilization through a multi-scale multi-task feature interaction module. DeMT~\cite{zhang2023demt} and DeMTG~\cite{xu2023deformable} further enhanced the integration of multi-task features through sophisticated task feature fusion and querying modules.
More recently, research has explored new paradigms beyond conventional interaction modeling, such as comprehensive bridge-based feature aggregation~\cite{zhang2025bridgenet} and diffusion-based multi-task learning frameworks~\cite{ye2024diffusionmtl,yangmulti}, which leverage generative denoising processes to jointly model task relationships and improve dense prediction performance.

Additionally, the Mixture of Experts (MoE) model has recently demonstrated impressive results in MTL by employing a group of experts to capture different aspects of data from various perspectives, thus creating distinct modalities~\cite{ye2023taskexpert, fan2022m3vit}. This approach uses a task-specific routing network to selectively fuse these modal data, facilitating targeted information sharing across tasks and enabling the modeling of complex task relationships. TaskExpert~\cite{ye2023taskexpert} leverages MLP experts for dynamic task routing, while M$^3$ViT~\cite{fan2022m3vit} integrates MoE layers into a vision transformer backbone. This integration sparsely activates task-specific experts during training, effectively disentangling the parameter space and minimizing conflicts between tasks, thereby enhancing the flexibility and efficiency of the learning process in intricate multi-task scenarios.

\begin{figure*}
    \centering
    \includegraphics[width=1\linewidth]{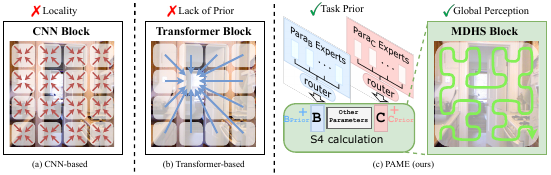}
    \caption{Comparison with other multi-task learning methods. 
    (a) CNN-based methods: These approaches are constrained by the local receptive fields of convolutions, hindering their ability to capture global context in multi-task scenarios, even with the inclusion of mixture of experts (MoE). (b) Transformer-based methods: While these methods can capture task-specific contexts, they lack inherent task priors. (c) Our PAME: By leveraging parameter experts, our method enables comprehensive parameter interactions across tasks and incorporates task priors to facilitate more effective task decoding.}
    \label{fig:enter-label}
\end{figure*}

Despite these advancements, multi-task learning still faces several challenges. CNN-based methods, known for their efficiency as shown in~\cite{gao2019nddr,vandenhende2020mti,ghiasi2021multi}, are limited by architectural constraints that restrict the global flow of information across tasks, as emphasized in~\cite{wang2018non,bello2019attention}. 
Moreover, although the incorporation of mixture of experts models into decoders has shown to be promise~\cite{ye2023taskexpert,yang2024multi}, CNNs still face significant challenges in handling complex multi-task scenarios as they tend to learn local parameters. Efforts to expand their capabilities, such as employing dilated convolutions~\cite{zhang2023rethinking} and scaled dot-product attention for context distillation~\cite{bruggemann2021exploring}, have yet to overcome issues with scalability. Transformer architectures are generally effective in many applications, but they struggle to capture task-specific priors in MTL through simple designs, limiting their capacity to distinguish tasks at a more granular level. Although Transformers possess global modeling capabilities, constructing Transformer expert models that effectively integrate these global modeling abilities remains challenging.

In response, we propose an innovative Parameter Aware Mamba Model (PAMM) based on the Mamba~\cite{gu2023mamba} framework, noted for its scalable parameters and linear computational efficiency in both training and inference.  This framework supports robust feature learning across varying spatial scales. Nonetheless, directly integrating Mamba's blocks can incur computational costs akin to those seen in conventional Transformer architectures. To address this, our MoE architecture incorporates the structured state space sequence (S4) model, optimizing global parameter computation and enhancing task-specific interactions. As depicted in Fig.~\ref{fig:overall architecture}(a), our overall architecture emphasizes multi-task and multi-scale perception within the decoder. 
Specifically, we apply convolution to refine the general features from the backbone for localized task-specific perception. The Parameter Aware Mamba Experts (PAME) module then extracts task parameters, enhancing task sharing and reducing conflicts. Within PAME module, a skip connection relays earlier task-specific local features to the task decoding head, synchronizing with global task decoding features from the PAME module to generate the final multi-task predictions. We have designed parameters for Mamba, establishing Parameter Experts (PE) and Parameter Prior (PP). PE models task relationships and learns task patterns by merging modal data generated by numerous experts through a learning fusion process. PP models the intrinsic properties of tasks. By integrating data from different modalities of parameters, we construct a comprehensive multi-task optimization process from multiple sub-aspects.
Additionally, we employ the Multi-Directional Hilbert Scanning (MDHS) method to enhance holistic feature collaboration, countering the directional sensitivity inherent to the causal Mamba model. 
The contributions can be summarized as follows:

\begin{itemize}
\item We introduce PAMM, an architecture combining Mamba and MoE within the model's decoder for multi-task dense prediction.
\item The proposed PAME module enhances the selection of hidden states by effectively harnessing MoE to balance the relations among task-specific parameters and leveraging S4 model to capture their long-range dependencies.
\item To enhance the perception of task characteristics, PAME incorporates task-specific priors. To mitigate the inherent mismatch of Mamba's unidirectional modeling characteristics with image data, a multi-directional scanning method is employed.
\item We validate the effectiveness of our design through extensive experiments on two challenging benchmarks, NYUD-v2 and PASCAL-Context. The results, both quantitative and qualitative, consistently affirm the robustness and superiority of our proposed approach.
\end{itemize}

Unlike CNN-based methods~\cite{xu2018pad,vandenhende2020mti}, whose local receptive fields hinder global consistency, and Transformer-based approaches~\cite{ye2022invpt,ye2024invpt++}, which lack explicit task priors, our PAMM framework introduces a parameter-aware paradigm for multi-task learning. By unifying state-space modeling, mixture-of-experts interactions, and multi-directional Hilbert scanning within a single decoder, it enables long-range context propagation, task-conditioned routing, and structure-aware aggregation. This design yields globally consistent, prior-guided, and task-adaptive dense predictions, addressing the representational limitations of previous architectures.

\section{Related Works}
\subsection{Mamba}
The emergence of Mamba~\cite{gu2023mamba}, a class of selective state space model, delivers a groundbreaking alternative for vision foundation models. Through selective scanning mechanism and hardware-aware algorithm, Mamba demonstrates superiority in capturing long-range contexts with linear computational costs compared to Transformer. 
Since Vim~\cite{zhu2024vision} and VMamba~\cite{liu2024vmamba} adapted Mamba in 2D non-causal images by employing multi-directional scanning strategies, a significant amount of researches have leveraged Mamba for various visual tasks to date.
For example, studies~\cite{yang2024vivim, yue2024medmamba, wu2025sk, SAM-Mamba, cheng2025mamba} have applied Mamba in medical image segmentation, achieving outcomes surpassing existing benchmarks. Some efforts~\cite{chen2024video, li2024videomamba, MambaAdapter, AVS-Mamba, xu2025th, zhao2025mamba, 11210145} investigate Mamba's potential for time series modeling. Some works~\cite{zhou2025state, jin2025unimamba} have utilized Mamba for perception in three-dimensional space.
Moreover, some researches~\cite{pioro2024moe,anthony2024blackmamba,lieber2024jamba} have concentrated on combining Mamba with MoE to validate the synergy of the two architectures and the scalability of Mamba in natural language processing fields. 
Inspired by these pioneers, we aim to delve into the capabilities of Mamba in multi-task dense prediction. Our research focuses on modulating the hidden state selective process of Mamba through MoE, thereby promoting joint optimization of multiple tasks.
\subsection{Multi-task Learning}
In contrast to single-task learning, which primarily explores novel network components and connectivity architectures, multi-task frameworks~\cite{Misra_Shrivastava_Gupta_Hebert_2016,mallya2018piggyback,sun2020learning,strezoski2019many,10086687,9337868,9740668,9970586,6963417,8638802,9781346,9262911} address multiple tasks by leveraging both task-specific and shared structures and parameters. These frameworks aim to enhance generalization through inter-task information sharing and minimize negative transfer~\cite{crawshaw2020multi}. 
Recent studies have further advanced this line of research from multiple perspectives. 
Some works~\cite{nishi2024joint} move beyond pairwise task relationship modeling by mapping all tasks into a unified latent space to capture holistic inter-task dependencies, while others~\cite{huang2024going} employ EM-based synergy embedding to explicitly model inter-task collaboration. 
Moreover, adaptive message passing methods~\cite{sirejiding2024adaptive} focus on local–global spatial interactions for dynamic cross-task communication, and efficiency-oriented approaches~\cite{shang2024efficient} introduce weight and activation binarization to significantly reduce computational and memory costs without sacrificing accuracy.
A common approach involves manually designing architectures that share model parameters across tasks. Additionally, some studies have employed techniques such as neural architecture search~\cite{zoph2016neural} or routing networks~\cite{rosenbaum2017routing} to learn cross-task sharing patterns and automatically derive architectures.

The dynamic nature of the mixture of experts (MoE) model makes it particularly well-suited for multi-task learning. By sharing a set of experts across different tasks, MoE enables joint optimization and effectively reduces the risk of negative transfer~\cite{crawshaw2020multi}. The Taskexpert framework~\cite{ye2023taskexpert} introduced a group of expert networks that decompose backbone features into task-general components, which are then transformed into task-specific features via dynamic task-specific gating networks. Similarly, MTLoRA~\cite{agiza2024mtlora} utilized task-agnostic and task-specific low-rank adapters to disentangle the parameter space during MTL fine-tuning, thereby enhancing the model's ability to manage task specialization and interaction within the MTL framework. Mod-squad~\cite{chen2023mod} integrated MoE layers into a transformer model, incorporating a novel loss function to address task and expert interdependencies. While existing methods~\cite{ye2023taskexpert,yang2024multi,agiza2024mtlora} have primarily focused on decoder components, their expert mechanisms, whether MLPs or CNNs, lack the ability to facilitate global task sharing. Our approach introduces task-specific expert interactions within the Mamba framework, leveraging a comprehensive state space for shared parameter modeling across multiple tasks.

\subsection{Multi-task Dense Prediction}
Multi-task dense prediction represents a significant advancement over single-task dense prediction approaches~\cite{chen2023kepsalinst,Liu2022DNA}. While single-task models focus on pixel-level predictions for a specific task, multi-task dense prediction aims to handle multiple related tasks simultaneously within a unified framework. This approach not only enhances training efficiency but also leverages shared features across tasks, potentially improving the generalization capabilities of the model.

\subsubsection{CNN-based Methods}
PAD-Net~\cite{xu2018pad} introduced an approach that combined preliminary predictions of depth estimation, scene parsing, surface normal estimation, and contour prediction to produce refined predictions for depth and scene parsing. This architecture was extended by PAP~\cite{zhang2019pattern} through the incorporation of an affinity learning layer, which learned pair-wise relations representing tasks and combined features from various tasks based on these relationships. MTI-Net~\cite{vandenhende2020mti} modeled task interactions across multiple scales of the receptive field. NDDR-CNN~\cite{gao2019nddr} utilized neural discriminative dimensionality reduction to enhance feature fusion operations in parallel layers. Cross-Stitch~\cite{Misra_Shrivastava_Gupta_Hebert_2016} offered separate networks for each task and facilitates information flow cross-talk between parallel layers in task networks, thereby enhancing the inter-relationship of task training. While these methods achieved notable results on mainstream benchmarks, they were inherently limited by the nature of CNNs, which could only capture local task features for sharing or interacting with task information. This limitation rendered them ineffective in scenarios requiring extensive task interactions. Thanks to the global modeling capabilities of the Mamba~\cite{gu2023mamba} framework, our method can effectively interact with tasks at the level of the entire feature map, thereby overcoming these limitations.
\subsubsection{Transformer-based Methods}
The Transformer architecture~\cite{vaswani2017attention} has substantially advanced artificial intelligence by providing superior global modeling capabilities and powerful pre-trained models that have notably improved the performance of Multi-Task Learning (MTL) methods. InvPT~\cite{ye2022invpt} was the pioneering work to employ the Transformer within a unified framework for simultaneous modeling of spatial positions and multiple tasks. Building on this, InvPT++~\cite{ye2024invpt++} enhanced the model by introducing a robust selective attention strategy in its cross-scale self-attention module, which significantly reduced redundancy in self-attention computations by leveraging cross-scale attention to boost performance through enhanced information interaction from previous layers.
MTFormer~\cite{xu2022mtformer} introduced a cross-task attention mechanism and a self-supervised cross-task contrastive learning algorithm to improve Multi-Task Learning (MTL) outcomes. MQTransformer~\cite{xu2023multi} was designed with a cross-task query attention module to infer dependencies among multiple task-related queries. DeMT~\cite{zhang2023demt} and DeMTG~\cite{xu2023deformable} combined the advantages of deformable CNNs and query-based Transformer with shared gating for dense prediction in multi-task learning. These methods focused on global-scale task interactions, leveraging the Transformer’s robust scalability to enable MTL models to learn global representations of different tasks, thereby significantly boosting the performance of MTL tasks. However, these approaches did not address the inherent intrinsic properties of different tasks, or task priors. Our approach incorporates task priors within the Mamba framework to thoroughly explore and leverage the intrinsic properties of tasks for improved MTL performance.

\begin{figure*}
    \centering
    \includegraphics[width=0.95\linewidth]{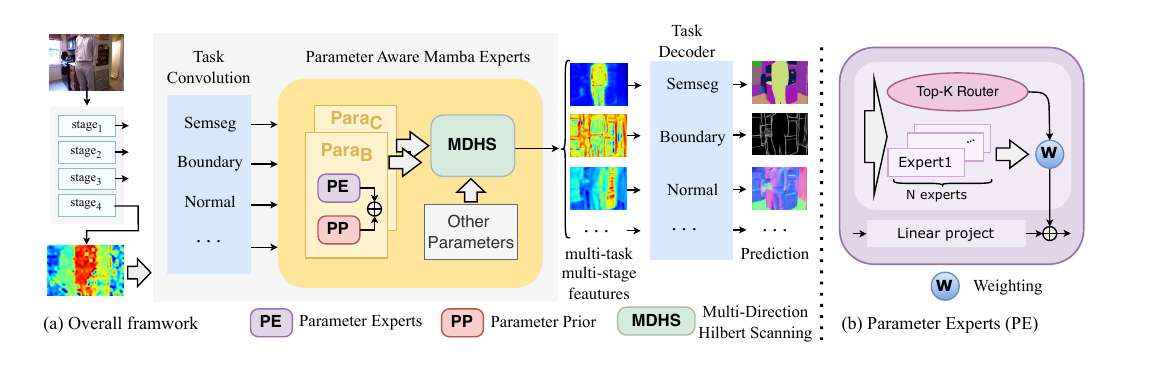}
    \caption{Overview of the architecture. (a) The process initiates with task-specific local feature extraction through task convolution. Following this, the Parameter Aware Mamba Experts (PAME) module configures task experts within Mamba's parameter space to promote global task interaction and integrates task priors for detail enhancement. Ultimately, it aggregates features from multiple scales for task-specific decoding. (b) The structure of Parameter Experts. To facilitate joint optimization across different tasks, we construct a task expert mixed network based on two parameters of Mamba. The multi-task experts are weighted and fused by a routing network, while also establishing non-shareable task-specific paths for different tasks.}
    \label{fig:overall architecture} 
\end{figure*}
\section{Preliminary}
\subsection{State Space Model.} State-space models (SSMs) originate in control theory and describe continuous, linear time-invariant systems. These models transform an one-dimensional signal $x(t)\in{\mathbb{R}^{L}}$ into a response $y(t)\in{\mathbb{R}^{L}}$ by mapping and transmission in the hidden state space $h(t)\in{\mathbb{R}^{N}}$, where $L$ and $N$ are denoted as the length of the input signal and the dimension of the hidden state space. The process can be formulated as linear ordinary differential equations (ODEs):
\begin{equation}
    h'(t)=\mathbf{A}h(t) + \mathbf{B}x(t),
\end{equation}
\begin{equation}
    y(t)=\mathbf{C}h(t) + \mathbf{D}x(t),
\end{equation}
where $\mathbf{A}\in{\mathbb{R}^{N\times{N}}}$ represents the evolution matrix, $\mathbf{B}\in{\mathbb{R}^{N}}$ and $\mathbf{C}\in{\mathbb{R}^{N}}$ are the projection matrix and $\mathbf{D}\in{\mathbb{R}^{1}}$ represents the skip connection. 

For deep learning scenarios, the continuous parameters $\textbf{A}$ and $\textbf{B}$ need to be discretized using zero-order hold (ZOH) techniques:
\begin{equation}\label{eq3}
        \mathbf{\bar{A}}=\text{exp}(\mathbf{\Delta{A}}), 
\end{equation}
\begin{equation}\label{eq4}
        \mathbf{\bar{B}}=(\text{exp}(\mathbf{\Delta{A}})-I)(\mathbf{\Delta{A}})^{-1}\mathbf{B},    
\end{equation}
\begin{equation}\label{eq5}
        \mathbf{\bar{C}}=\mathbf{C},
\end{equation}
where $\mathbf{\Delta}$ denotes the timescale parameter, and $\mathbf{\bar{A}}$ and $\mathbf{\bar{B}}$ denote the discrete equivalents of the continuous parameters $\mathbf{A}$ and $\mathbf{B}$. 
The discretized formulation of SSM can be rewritten as follows:
\begin{equation}\label{eq6}
h_{k}=\mathbf{\bar{A}}h_{k-1}+\mathbf{\bar{B}}x_{k},
\end{equation}
\begin{equation}\label{eq7}
y_{k}=\mathbf{\bar{C}}h_{k}+\mathbf{\bar{D}}x_{k}
\end{equation}

Differing from the time-invariant nature of previous SSMs~\cite{gu2021efficiently,fu2022hungry}, Mamba~\cite{gu2023mamba} introduced a selective scanning mechanism, ensuring all parameters dependent on the input. 
To be specific, the parameters $\mathbf{B}\in{\mathbb{R}^{{L}\times{N}}}$, $\mathbf{C}\in{\mathbb{R}^{{L}\times{N}}}$ and $\mathbf{\Delta}\in{\mathbb{R}^{{L}\times{D}}}$ are projected by the input $x\in{\mathbb{R}^{{L}\times{D}}}$, where $D$ is the embeddings of $x$. Each parameter is integral to the functioning of the selective mechanism. For instance, $\mathbf{\Delta}$ dictates which information the model should focuses on or disregards. $\mathbf{B}$ allows fine-grained control over the encoding process from input to hidden state while $\mathbf{C}$ decodes the implicit context into the output.
\subsection{Mixture of Experts}
In multi-task learning, a mixture of experts model typically comprises two main components: an expert group \( E = \{E_1, E_2, \ldots, E_N\} \) with \( N \) experts, and a series of task gated networks \( \mathcal{G} = \{\mathcal{G}^1, \mathcal{G}^2, \ldots, \mathcal{G}^T\} \) for \( T \) tasks, while \( N \) weights are contained in every task's gate. For the feature \( x \), each expert provides a discriminative outcome \( y_j = E_j(x) \). These outcomes are then weighted by the task gated network, producing the output features \( y \), which can be expressed as:
\begin{equation}
\begin{split}
        y=\sum_{j=1}^{N}\mathcal{G}_j(x){E_j(x) }
\end{split}
\end{equation}

To balance the load among experts and enhance training stability, a common method employed is the use of noisy top-k routing~\cite{shazeer2017outrageously}. Specifically, during training, random noise is added to the output of all gates. Subsequently, only the top-k gates with the highest values are utilized, while the others are set to zero. The selected gates are then normalized using the softmax function:
\begin{equation}
\begin{split}
        \mathcal{R}(x) = \text{Top-K}\left(\text{Softmax}\left(\mathcal{G}(x) + N(0, 1) \cdot W_{\text{noise}}(x)\right)\right),
\end{split}
\end{equation}
where $\text{Top-K}(\cdot, k)$ sets all elements in the vector to zero except the elements with the largest K values,  \( N(0,1) \) denotes a noise sample drawn from a normal distribution with a mean of zero and a standard deviation of one, and \( W_{\text{noise}}(\cdot) \) calculates the corresponding noise weights. Thus, the output for each task can be expressed as:
\begin{equation}
\begin{split}
        y = \sum_{j=1}^{N}\mathcal{R}_j(x)E_j(x)
\end{split}
\end{equation}
\section{Methods}
\subsection{Overall Framework}
The architecture of our proposed Parameter Aware Mamba Model (PAMM) is depicted in Fig.~\ref{fig:overall architecture}(a). We utilize a pre-trained ViT~\cite{dosovitskiy2020image} as the backbone to extract multi-stage features. The input image is passed through this backbone, generating a set of feature tokens at each selected output stage. A task-specific convolution layer is then applied within the Task Convolution module to locally decode these features for the respective tasks. The resulting multi-task features are fed into the Parameter Aware Mamba Experts (PAME) module, where information is decomposed and fused and prior information is integrated, enabling global task interactions and comprehensive prior modeling. To address the potential loss of local information in holistic perception, skip connections~\cite{he2016deep} are employed within the PAME to transfer the locally decoded features from the Task Convolution to the Task Decoder. The Task Decoder then aggregates outputs from various layers to produce the final multi-task prediction map. The detailed workings of the proposed PAME will be elaborated in the following subsection.

\subsection{Parameter Aware Mamba Experts Module}
In this subsection, we discuss the detailed structure of our key component, the Parameter Aware Mamba Experts (PAME) module.  Equation (\ref{eq3}), (\ref{eq4}) and (\ref{eq5}) provide insight into the roles of these parameters:
\begin{itemize}
  \item Parameter $\mathbf{A}$ compresses past memories, aiming to reduce redundancy and focusing on relevant information.
  \item Parameter $\mathbf{D}$ serves as a skip connection, facilitating direct pathways between layers in deep networks and mitigating issues related to gradient flow.
  \item Parameter $\mathbf{\Delta}$ is crucial for maintaining the stability of the process, ensuring that the dynamic system does not diverge.
  \item Parameter $\mathbf{B}$ finely controls the impact of current inputs on the encoding of the hidden state, thus modulating the influence of new information.
  \item Parameter $\mathbf{C}$ decodes the final output from the hidden state space, which contains contextual memory from previous inputs.
\end{itemize}
\begin{figure}
    \centering
    \includegraphics[width=1.0\linewidth]{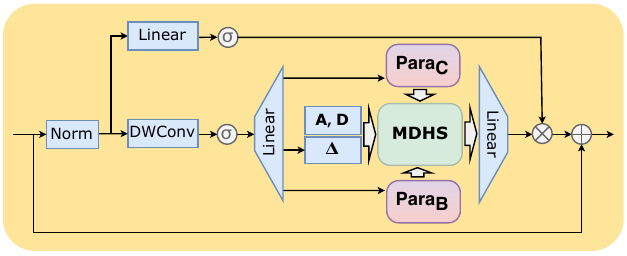}
    \caption{Details of the proposed Parameter Aware Mamba Experts (PAME) module. Building on established Mamba-based methods, PAME integrates depth-wise separable convolutions (DWConv)~\cite{sifre2014rigid}, gating, and skip connections. Key innovations include Parameter Experts and Parameter Priors for optimizing parameters B and C, along with a state space computational approach using Multi-Directional Hilbert Scannig (MDHS).
   }
    \label{fig:PAME}
\end{figure}
Specifically, parameters $\mathbf{B}$ and $\mathbf{C}$ customize transformations between sequences and state spaces for specific tasks, enabling selective sharing that avoids negative transfer~\cite{crawshaw2020multi}. Optimizing them with task-specific priors exploits complementary information across tasks, thereby improving both performance and adaptability.

The PAME module, depicted in Fig.~\ref{fig:PAME}, is consistent with the methodologies of established Mamba-based methods such as those proposed in~\cite{liu2024vmamba,zhu2024vision}, while incorporating several enhancements. These improvements include refined parameter calculations and advanced scanning techniques. Initially, local information captured by DWConv undergoes a dimensional expansion via a channel expansion layer, preparing it for state space parameter computation. In PAME, these parameters are derived from both a MoE and task-specific priors. The Multi-Directional Hilbert Scanning (MDHS) technique introduces a novel approach for converting 2D data into multi-directional 1D sequences, enabling more efficient state space computations. This process is completed by a linear layer that restores the original channel dimensions. Within PAME, gate networks and skip connections are integral. Skip connections facilitate the transfer of previously decoded local features from task-specific convolutions, while gate networks selectively regulate the processing of input data.
\subsubsection{Parameter Experts Calculation}
As previously discussed, parameters $\mathbf{B}$ and $\mathbf{C}$ are crucial for identifying and capturing distinct feature sequence patterns specific to individual tasks.  Our approach enhances each task's perceptual capabilities by effectively managing the interaction of shared patterns. The mixture of experts architecture, known for its significant impact in multi-task learning (MTL), facilitates the sharing of network structures based on input features. We leverage the MoE structure to obtain multiple constituent modals of parameters $\mathbf{B}$ and $\mathbf{C}$, and selectively fuse these modals to derive the fused parameter features. By constructing Parameter Experts (PE), our model not only facilitates joint optimization across tasks but also achieves more refined control of information flow compared to using numerous individual Mamba block experts. 

In the original vision Mamba block~\cite{zhu2024vision,liu2024vmamba}, parameters $\mathbf{B}$ and $\mathbf{C}$ are obtained through simple linear projection. To facilitate joint optimization across tasks, we have added a MoE pathway, as shown in Fig.~\ref{fig:overall architecture}(b). 
Specifically, each task possesses its own task routing network, which determines the preference weights for experts specific to different tasks. For input feature $x^i \in \mathbb{R}^{C\times H \times W}$, our task routing network is divided into two aspects: channel and spatial. On one hand, in the channel dimension, a linear projection $\text{Linear}_1$ reduces the input feature channels to one-quarter of its original size, followed by a global pooling to reshape it to $\mathbb{R}^{C/4\times 1}$. On the other hand, to fully capture global characteristics, we first apply global pooling to obtain features shaped $\mathbb{R}^{C\times 1}$, which are then transformed to $\mathbb{R}^{C/4 \times 1}$ through a linear projection $\text{Linear}_2$. Then, features from both aspects are concatenated and processed through a linear projection $\text{Linear}_3$ and a nonlinear layer $\text{Act}$, followed by a softmax function to obtain the final expert weights $\mathcal{G}$, as denoted below:
\begin{equation}
        f_r^i = \text{Cat}\left( \text{Pool}\left(\text{Linear}_1\left( x^i\right) \right) , \text{Linear}_2\left(\text{Pool}(x^i)\right)\right),
\end{equation}
\begin{equation}
        \mathcal{G}^i = \text{Softmax}\left(\text{Act}\left( \text{Linear}_3\left(f_r^i\right) \right) \right)     
\end{equation}

Subsequently, the top $K$ experts with the highest activation values are selected. The input task features $x$ are then fed to all experts, and their outputs are weighted according to the weights of the top-k experts. During training, Gaussian noise is added to enhance the stability of the network, as expressed in Equation (\ref{noise}). The final result $y_{moe}^i \in \mathbb{R}^{D \times H \times W}$ for each task is calculated using Equation (\ref{eq moe}). 
\begin{equation}\label{noise}
        \mathcal{R}^i = \text{Top-K}\left(\text{Softmax}\left(\mathcal{G}(x^i) + N(0, 1) \cdot W_{\text{noise}}(x^i)\right)\right),
\end{equation}
\begin{equation}\label{eq moe}
        y_{moe}^i = \sum_{j=1}^{N}\mathcal{R}_j^i(x^i){E_j(x^i) }
\end{equation}

To better capture the characteristics of individual tasks, we retain the original linear projection layer, which projects the channel dimension to the state space dimension $D$. 
In summary, the computation of PE can be expressed as follows:
\begin{equation}
    y_{pe}^i = y_{moe}^i + y_{linear}^i
\end{equation}
\subsubsection{Parameter Priors Calculation}
In our framework, task-specific priors are embedded within the features. We integrate these priors into the state-space model through distinct parameters, $\mathbf{B}$ and $\mathbf{C}$, termed as Parameter Priors (PP). Each task is associated with a unique set of learnable prior parameters, tailored to modify $\mathbf{B}$ and $\mathbf{C}$ accordingly. A prior parameter, $y_{p}^i \in \mathbb{R}^{D}$, which is spatially invariant, is defined and subsequently expanded to align with the spatial dimensions of the task-specific features. 
Denoting this prior as $y_{pp}^i$, the values for parameters $\mathbf{B}$ and $\mathbf{C}$ are finally calculated as:
\begin{equation}
    y_{final}^i = y_{pe}^i + y_{pp}^i
\end{equation}

\begin{figure}
    \centering
    \includegraphics[width=1\linewidth]{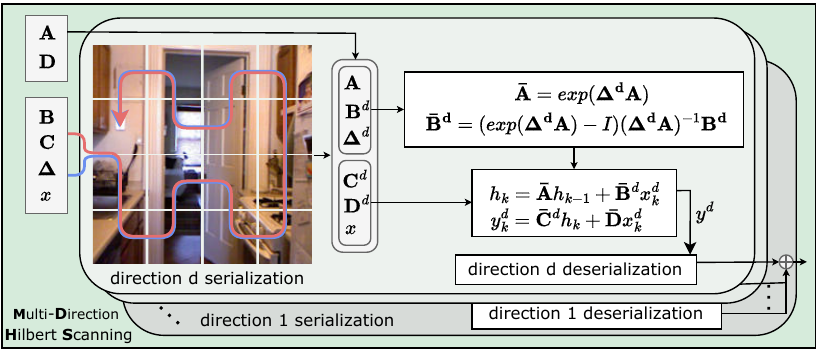}
    \caption{The proposed Multi-Directional Hilbert Scanning (MDHS) method serializes portions of the input parameters and the input $x$ following the Hilbert scanning approach. The output is then computed using a state equation, after which the original image order is restored through deserialization. Finally, the outputs from the scans in different directions are aggregated to produce the final result.}
    \label{fig:MHDS}
    \vspace{-3mm}
\end{figure}

\def\thickhline{\noalign{\hrule height 1pt}}
\begin{table*}[htbp]
    \setlength{\tabcolsep}{3pt}
    \caption{Comparison of different methods on the PASCAL-Context and NYUD-v2 datasets. \textbf{Bold} indicates the best performance for each metric when using the same backbone, while \underline{underlined} denotes the second-best. Arrows ($\uparrow$/$\downarrow$) signify that higher or lower values are preferable, respectively.}
    \label{tab:methods comparision}
    \vspace{2mm}
    \centering
    \scalebox{0.88}
    {\renewcommand{\arraystretch}{1.7}
    \begin{tabular}{
    >{\centering\arraybackslash}p{2.5cm}
    >{\centering\arraybackslash}p{2cm}
    >{\centering\arraybackslash}p{1.3cm}
    >{\centering\arraybackslash}p{1.3cm}
    >{\centering\arraybackslash}p{1.3cm}
    >{\centering\arraybackslash}p{1.3cm}
    >{\centering\arraybackslash}p{1.3cm}
    >{\centering\arraybackslash}p{1.3cm}
    >{\centering\arraybackslash}p{1.3cm}
    >{\centering\arraybackslash}p{1.3cm}
    >{\centering\arraybackslash}p{1.3cm}
    >{\centering\arraybackslash}p{1.3cm}
    >{\centering\arraybackslash}p{1.3cm}}
        \thickhline
        \multirow{3}{*}[0.1cm]{\textbf{Method}}  & \multirow{3}{*}[0.1cm]{\textbf{Backbone}} & \multicolumn{6}{|c|}{\textbf{PASCAL-Context}} & \multicolumn{5}{c}{\textbf{NYUD-v2}}\\
        \cline{3-13}
        & & \multicolumn{1}{|c}{\textbf{Semseg}} & \textbf{Parsing} & \textbf{Saliency} & \textbf{Normal} & \textbf{Boundary} &\multicolumn{1}{c|}{\textbf{$\Delta_{g}$}} & \textbf{Semseg} & \textbf{Depth} & \textbf{Normal} & \textbf{Boundary} &\textbf{$\Delta_{g}$}\\[-1.8ex]
        & & \multicolumn{1}{|c}{mIoU$\uparrow$} & mIoU$\uparrow$ & maxF$\uparrow$ & mErr$\downarrow$ & odsF$\uparrow$  &\multicolumn{1}{c|}{\%$\uparrow$} & mIoU$\uparrow$ & RMSE$\downarrow$ & mErr$\downarrow$ & odsF$\uparrow$  &\%$\uparrow$ \\
        \hline 

        InvPT~\cite{ye2022invpt}& Swin-L& \multicolumn{1}{|c}{\underline{78.53}}& \underline{68.58}& \underline{83.71}& \underline{14.56}&\underline{73.60} & \multicolumn{1}{c|}{\underline{-0.04}} & 51.76& \underline{0.5020}& \underline{19.39}& 77.60 &2.02\\[-1.5ex]
        InvPT++~\cite{ye2024invpt++}& Swin-L& \multicolumn{1}{|c}{\textbf{79.65}}& \textbf{69.14}& \textbf{84.78}& \textbf{14.09}& \textbf{74.80}& \multicolumn{1}{c|}{\textbf{1.68}}& \underline{52.35}& \textbf{0.4921}& \textbf{18.99}& \underline{77.90}& \textbf{3.36}\\[-1.5ex]
        MQTransformer~\cite{xu2023multi}& Swin-L& \multicolumn{1}{|c}{-}& -& -& -&- & \multicolumn{1}{c|}{-} & \textbf{54.84}& 0.5325& 19.67& \textbf{78.20} & \underline{3.04}\\
        \hline 
        M$^3$VIT~\cite{fan2022m3vit}& ViT-B& \multicolumn{1}{|c}{75.20}& 64.50& 66.1 &14.80& \textbf{72.60} & \multicolumn{1}{c|}{-6.90} & 49.10& 0.5570& -& - & -1.65\\ [-1.5ex]

        InvPT~\cite{ye2022invpt}& ViT-B& \multicolumn{1}{|c}{\underline{77.33}}& 66.62& 85.14& 13.78&73.20 & \multicolumn{1}{c|}{0.42} & \underline{50.30}& 0.5367& \underline{19.04}& \textbf{78.10} &\underline{0.42}\\[-1.5ex]
        InvPT++~\cite{ye2024invpt++}& ViT-B& \multicolumn{1}{|c}{76.95}& \underline{66.89}& \underline{85.12} &\underline{13.54}& 73.30 & \multicolumn{1}{c|}{\underline{0.77}} & 49.79& \textbf{0.5318}& \textbf{18.90}& 77.10 & 0.25\\[-1.5ex]
        PAME (ours)& ViT-B& \multicolumn{1}{|c}{\textbf{79.12}}& \textbf{67.83}&\textbf{85.25} &\textbf{13.40}& \underline{73.40} & \multicolumn{1}{c|}{\textbf{1.86}} & \textbf{51.10}& \underline{0.5349}& \textbf{18.90}& \underline{78.00} & \textbf{1.02}\\
        \hline 
        Baseline& ViT-L& \multicolumn{1}{|c}{78.67}& 67.75& 84.13& 14.01& 71.32&\multicolumn{1}{c|}{0.00} & 53.36& 0.5635& 19.25& 76.90&0.00\\[-1.5ex]
        InvPT~\cite{ye2022invpt}& ViT-L& \multicolumn{1}{|c}{79.03}& 67.61& 84.81& 14.15&73.00 & \multicolumn{1}{c|}{0.48} & 53.56& 0.5183& 19.00& 77.60 &2.65\\[-1.5ex]
        InvPT++~\cite{ye2024invpt++}& ViT-L& \multicolumn{1}{|c}{80.22}& 69.12& 84.74& 13.73& \underline{74.20}& \multicolumn{1}{c|}{2.15}& 53.85& \textbf{0.5096}& 18.67& 78.10& 3.76\\[-1.5ex]
        TaskPrompter~\cite{ye2022taskprompter}& ViT-L& \multicolumn{1}{|c}{\underline{80.89}}& 68.89& \underline{84.83}& 13.72& 73.50& \multicolumn{1}{c|}{2.09} & 55.30& 0.5152& \textbf{18.47}& \underline{78.20} &\underline{4.49}\\[-1.5ex]
        TaskExpert~\cite{ye2023taskexpert}& ViT-L& \multicolumn{1}{|c}{80.64}& \underline{69.42}& \textbf{84.87}& \underline{13.56}& 73.30& \multicolumn{1}{c|}{\underline{2.37}} & \underline{55.35}& 0.5157& 18.54& \textbf{78.40}&4.46\\[-1.5ex]
        PAME (ours)& ViT-L& \multicolumn{1}{|c}{\textbf{81.38}}& \textbf{70.39}& 84.51& \textbf{13.36}& \textbf{75.30} & \multicolumn{1}{c|}{\textbf{3.60}} & \textbf{56.63}& \underline{0.5102}& \underline{18.52}& 78.10 & \textbf{5.23}\\
        \thickhline
    \end{tabular}}
\end{table*}

\subsubsection{Hilbert Scanning Mechanism}
Previous studies have explored various scanning directions~\cite{liu2024vmamba,zhu2024vision}. We leverage the Hilbert scanning (H-scan) mechanism and construct a multi-directional method, Multi-Directional Hilbert Scanning (MDHS), as shown in Fig.~\ref{fig:MHDS}. The H-scan, recognized for its effective sequential causalization of point clouds~\cite{wu2023point}, has been further developed in our study to include four directional scans for enhanced image and parameter serialization. To accommodate various spatial dimensions of feature maps, we construct the smallest Hilbert curve larger than the feature shape and adjust through 90-degree rotations to establish multiple scanning directions. For feature maps smaller than the designed curve, we tailor the 2D to 1D serialization by cropping from the bottom-left, ensuring accurate alignment with the original image topology before summation to derive the final H-scan outputs $x^d$ $(d=1,2,3,4)$. The image serialization function can be formalized as:
\begin{equation}
x^d = \text{Serial}^d(x),
\end{equation}
where $\text{Serial}^d$ refers to the d-th method of feature serialization.
Parameters $\mathbf{B}$, $\mathbf{C}$ and $\mathbf{\Delta}$ are all serialized in four directions as the same as the input feature $x$, and S4 calculations are then performed for each direction, as expressed in Equation (\ref{eq s4_1}) and Equation (\ref{eq s4_2}). Finally, the results from the four directions are deserialized to restore the 1D features back to the shape of the original 2D input features, and these are summed to produce the final output of MDHS, as expressed in Equation (\ref{eq sum}):
\begin{equation}\label{eq s4_1}
    h_{k}=\mathbf{\bar{A}}h_{k-1}+\mathbf{\bar{B}}^dx_{k}^d,
\end{equation}
\begin{equation}\label{eq s4_2}
    y_{k}^d=\mathbf{\bar{C}}^dh_{k}+\mathbf{\bar{D}}x_{k}^d,
\end{equation}
\begin{equation}\label{eq sum}
    y=\sum_{d=1}^{4}\text{Deserial}^d(y^d),
\end{equation}
where $\text{Deserial}^d$ represents the inverse operation of the corresponding d-th serialization. 

\subsection{Task Decoder}
The Task Decoder generates the final predictions by fusing the outputs from multi-level PAME, as illustrated in Fig.~\ref{fig:overall architecture}(a). This involves concatenating the outputs $F_s^i \in \mathbb{R}^{C\times H\times W}$ from each PAME stage along the channel dimension. A multi-layer perceptron (MLP) is then employed to project these concatenated channels to the number of stages. We subsequently apply a softmax function to calculate channel-wise weights, denoted by $W_s^i \in \mathbb{R}^{S \times H \times W}$, representing the weights for the $S$ stages. These weighted outputs are summed to produce the final task feature, which is decoded into the prediction using a convolutional head. The sequential operations are defined by the following operations:
\begin{equation}
\begin{split}
    W_s^i = \text{Softmax}\left(\text{MLP}\left(\text{Concat}\left(F_1^i,...,F_S^i\right)\right)\right),
\end{split}
\end{equation}
\begin{equation}
    TF^i = \sum_{s=1}^{S}W_s^i\times F_s^i,
\end{equation}
\begin{equation}
    Pre^i = \text{ConvHead}^i(TF^i),
\end{equation}
where $TF$ and $Pre$ mean task feature and final prediction result, $\text{Concat}(\cdot)$ denotes the concatenation of feature maps across channels, $\text{MLP}(\cdot)$ involves two layers of linear projections, and $\text{ConvHead}(\cdot)$ is a convolutional layer tailored to match the number of channels in the ground truth of corresponding task.
\begin{figure}[t]
    \centering 
    \includegraphics[width=1\linewidth]{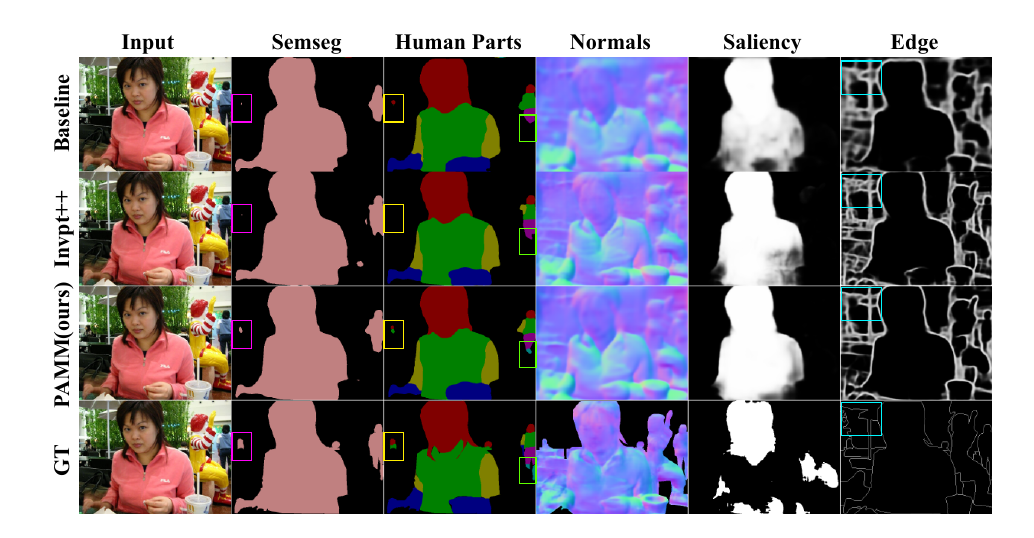}
    \caption{Visual comparison of multi-task predictions generated by our method and InvPT++ on the PASCAL-Context dataset. Our method exhibits enhanced generalization capabilities, enabling better capture of features of small objects.}
    \label{fig:pascal_vis} 
\end{figure} 
\section{Experiments}
\subsection{Experimental Settings}
\subsubsection{Datasets} We evaluate our multi-task learning (MTL) model on two well-established benchmarks: NYUD-v2~\cite{silberman2012indoor} and PASCAL-Context~\cite{chen2014detect}. The NYUD-v2 dataset, which includes comprehensive annotations for various tasks such as semantic segmentation, monocular depth estimation, surface normals, and object boundary detection, features 795 training images and 654 test images. This dataset provides a robust framework for assessing the effectiveness of our model across multiple tasks. Similarly, PASCAL-Context, with its diverse annotations covering semantic segmentation, human parsing, saliency detection, surface normals, and object boundary detection, includes 4998 training images and 5105 test images. This larger set allows us to evaluate the scalability and robustness of our MTL approach under varied and complex scenarios. 

\begin{figure}[t]
    \centering  
    \includegraphics[width=1.0\linewidth]{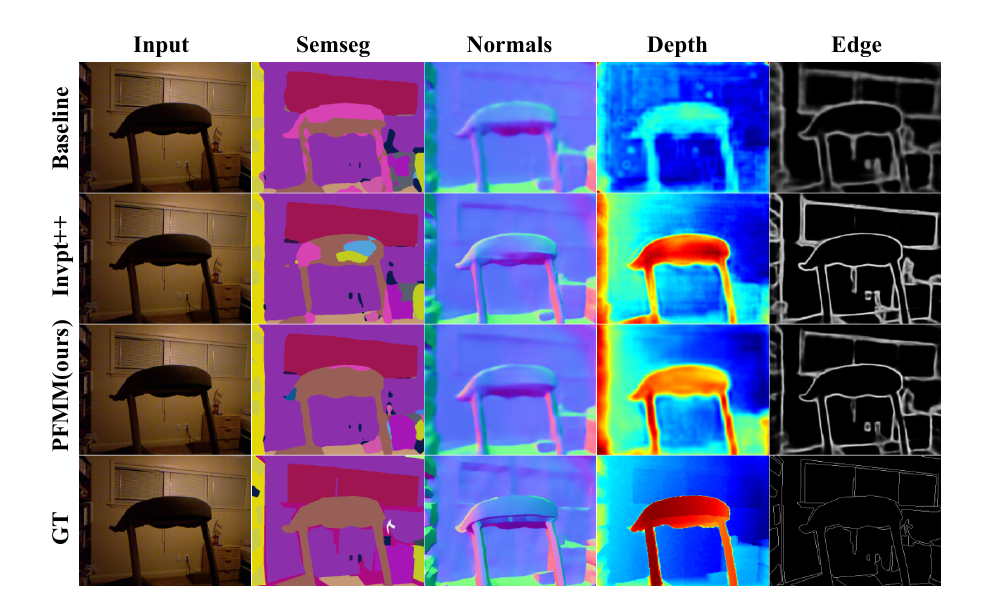}
    \caption{Visual comparison of multi-task predictions generated by our method and InvPT++ on the NYUD-v2 dataset. It is evident that our method produces predictions with significant improvements, resulting in more consistent outcomes across extensive areas of prediction.}
    \label{fig:nyu_vis}
\end{figure}
\subsubsection{Evaluation Metrics} 
Our evaluation follows established benchmarks for multi-task learning in dense prediction scenarios. In line with previous works ~\cite{ye2022invpt}, we assess semantic segmentation (Semseg) and human parsing (Parsing) using mean intersection-over-union (mIoU). Surface normal estimation (Normal) is evaluated based on the mean error (mErr), while monocular depth estimation (Depth) is assessed using root mean square error (RMSE). Saliency detection (Saliency) is measured by the maximal F-measure (maxF), and object boundary detection (Boundary) is evaluated using the optimal-dataset-scale F-measure (odsF). To provide a comprehensive performance assessment across all tasks, we calculate the multi-task performance as the average gain per task relative to our baseline, denoted as $\Delta{_g}$.
\subsubsection{Implementation Details} 
We employed the ViT (ViT-B and ViT-L) as the backbone to benchmark against other methods. For ablation analyses, we configured the ViT-B as the foundational backbone. We set top-k=9 out of 15 experts in PE. Our multi-task learning network undergoes training across 40,000 iterations with a batch size of 6 on both datasets. We maintained the identical optimizer and loss functions as specified in InvPT~\cite{ye2022invpt}. Our comprehensive baseline incorporates both the backbone and the Task Decoder, and we apply a unified multi-task loss to supervise the network.

\begin{figure*}
    \centering
    \includegraphics[width=1.05\linewidth]{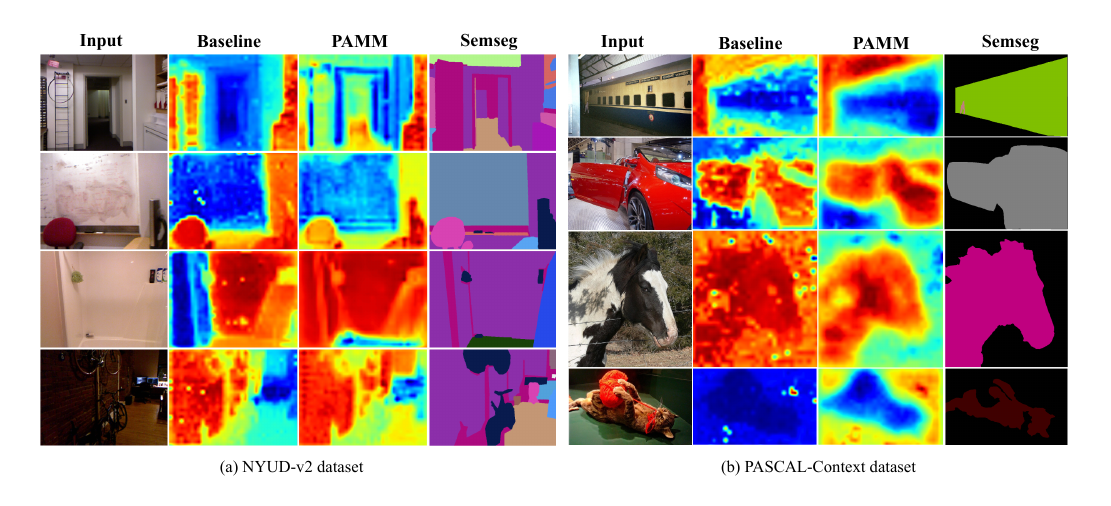}
    \caption{PCA visualization comparing semantic segmentation features between our method and the baseline on (a) the NYUD-v2 dataset and (b) the PASCAL-Context dataset. We performed PCA dimensionality reduction and visualized the semantic segmentation features from the final decoding head. By referencing the ground truth semantic masks, our method achieves greater semantic consistency and more distinct separation between different semantic classes.}
    \label{fig:pca} 
\end{figure*}
\begin{table}[tp]
  \caption{Ablation study of the main components of our method on the PASCAL-Context dataset, including Task Convolution (TC) and parameter aware mamba experts (PAME) module.}
  \label{tab:ablation main block}
    
    \vspace{2mm}
    \centering
    \scalebox{0.95}{
        \renewcommand{\arraystretch}{1.7}
        \begin{tabular}{
        >{\centering\arraybackslash}p{0.3cm}
        >{\centering\arraybackslash}p{0.7cm}
        >{\centering\arraybackslash}p{0.6cm}
        >{\centering\arraybackslash}p{0.7cm}
        >{\centering\arraybackslash}p{0.7cm}
        >{\centering\arraybackslash}p{0.7cm}
        >{\centering\arraybackslash}p{0.8cm}
        >{\centering\arraybackslash}p{0.8cm}}
        \thickhline
        \multicolumn{2}{c}{\hspace{0cm}\textbf{Module}} & \multicolumn{1}{|c}{\textbf{Semseg}} & \textbf{Parsing} & \textbf{Saliency} & \textbf{Normal} & \textbf{Boundary} &\textbf{$\Delta_{g}$} \\[-1.5ex]
        \multirow{1}{*}[-0.1cm]{\textbf{TC}}& \multirow{1}{*}[-0.1cm]{\textbf{PAME}} & \multicolumn{1}{|c}{mIoU$\uparrow$} & mIoU$\uparrow$ & maxF$\uparrow$ & mErr$\downarrow$ & odsF$\uparrow$  &\%$\uparrow$ \\
         
        \hline
        \ding{55}& \ding{55}  & \multicolumn{1}{|c}{77.12} & 66.16 & 84.71 & 13.80 & 69.80 & 0.00 \\[-1.5ex]

        \checkmark &\ding{55} &  \multicolumn{1}{|c}{\underline{78.01}}& \underline{67.35} & \underline{84.92} & \underline{13.59}& \underline{70.10}& \underline{1.03}\\[-1.5ex]
        
        \checkmark& \checkmark & \multicolumn{1}{|c}{\textbf{79.12}}& \textbf{67.83}& \textbf{85.25}& \textbf{13.40}& \textbf{73.40}&\textbf{2.76}\\
        
        \thickhline
        
  \end{tabular}}
\end{table}

\begin{table}[tp]
    \setlength{\tabcolsep}{3pt}
    \caption{Ablation study of Parameter Experts (PE) on the PASCAL-Context dataset, including two parameters}
    \label{tab:ablation PE}
    
    \vspace{2mm}
    \centering
    \scalebox{0.98}{
        \renewcommand{\arraystretch}{1.7}
        \begin{tabular}{
        >{\centering\arraybackslash}p{0.6cm}
        >{\centering\arraybackslash}p{0.6cm}
        >{\centering\arraybackslash}p{1cm}
        >{\centering\arraybackslash}p{1cm}
        >{\centering\arraybackslash}p{1cm}
        >{\centering\arraybackslash}p{1cm}
        >{\centering\arraybackslash}p{1cm}
        >{\centering\arraybackslash}p{1cm}}
        \thickhline
        \multicolumn{2}{c}{\textbf{PE}} & \multicolumn{1}{|c}{\textbf{Semseg}} & \textbf{Parsing} & \textbf{Saliency} & \textbf{Normal} & \textbf{Boundary} &\textbf{$\Delta_{g}$}\\[-1.8ex]
        $\textbf{B}$ & $\textbf{C}$ & \multicolumn{1}{|c}{mIoU$\uparrow$} & mIoU$\uparrow$ & maxF$\uparrow$ & mErr$\downarrow$ & odsF$\uparrow$  & \%$\uparrow$ \\
        \hline
        $\checkmark$ & $\checkmark$ & \multicolumn{1}{|c}{\textbf{79.12}}& \textbf{67.83}& \textbf{85.25}& \underline{13.40} &  \textbf{73.40} & \textbf{0.52}\\[-1.5ex]
        \ding{55}& $\checkmark$ & \multicolumn{1}{|c}{\underline{78.94}}& \underline{67.63}& 84.90& \textbf{13.37}& \underline{73.20}&\underline{0.32}\\[-1.5ex]
        $\checkmark$ &\ding{55}& \multicolumn{1}{|c}{78.95}& 67.56& 84.81& 13.43& \underline{73.20}&0.19\\[-1.5ex]
        \ding{55}&\ding{55}& \multicolumn{1}{|c}{78.74}& 67.42& 85.01& 13.49& 73.00&0.00\\
        \thickhline
    \end{tabular}}
\end{table}

\subsection{Comparisons with Other Methods}
\subsubsection{Quantative Results Analysis}
We conducted a fair comparison using Vision Transformer~\cite{dosovitskiy2020image} as the backbone across two widely recognized public datasets, ensuring the same architecture and evaluation metrics as in prior studies. Our approach was benchmarked against InvPT~\cite{ye2022invpt}, InvPT++~\cite{ye2024invpt++}, TaskPrompter~\cite{ye2022taskprompter}, M$^3$ViT~\cite{fan2022m3vit}, MQTransformer~\cite{xu2023multi}, and TaskExpert~\cite{ye2023taskexpert}. As shown in TABLE~\ref{tab:methods comparision}, our method achieved higher $\Delta{_g}$ values than all previous methods on both datasets. We also compared our approach against methods that employed the Swin Transformer~\cite{liu2021swin}, a hierarchical model widely recognized for its effectiveness in dense prediction tasks~\cite{ye2022invpt,ye2024invpt++,xu2023multi}. Our method consistently outperformed these alternatives in terms of $\Delta{_g}$ values. These findings underscore the effectiveness of our approach and demonstrate the practical advantages of integrating state-space models into multi-task learning.
\subsubsection{Qualitative Results Analysis}
\begin{figure}[t]
    \centering 
    \includegraphics[width=1\linewidth]{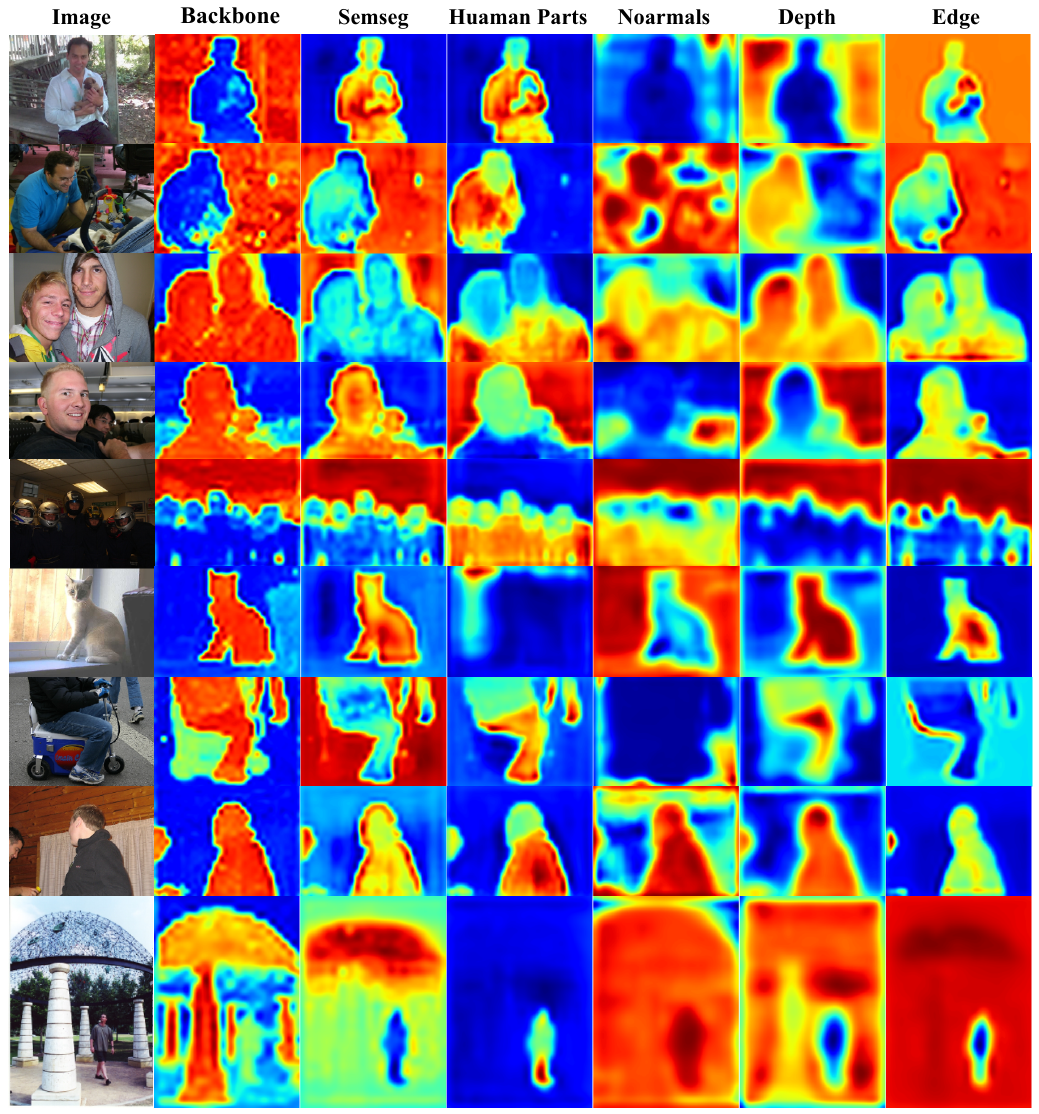}
    \caption{The feature visualization comparison on PASCAL-Context. From left to right are the input image, the PCA visualization of the general feature from the final layer of the backbone, and the PCA visualizations of the five task-specific features refined by the PAME module.}
    \label{fig:pascal_self_pca} 
\end{figure}

\begin{figure}[t]
    \centering 
    \includegraphics[width=1\linewidth]{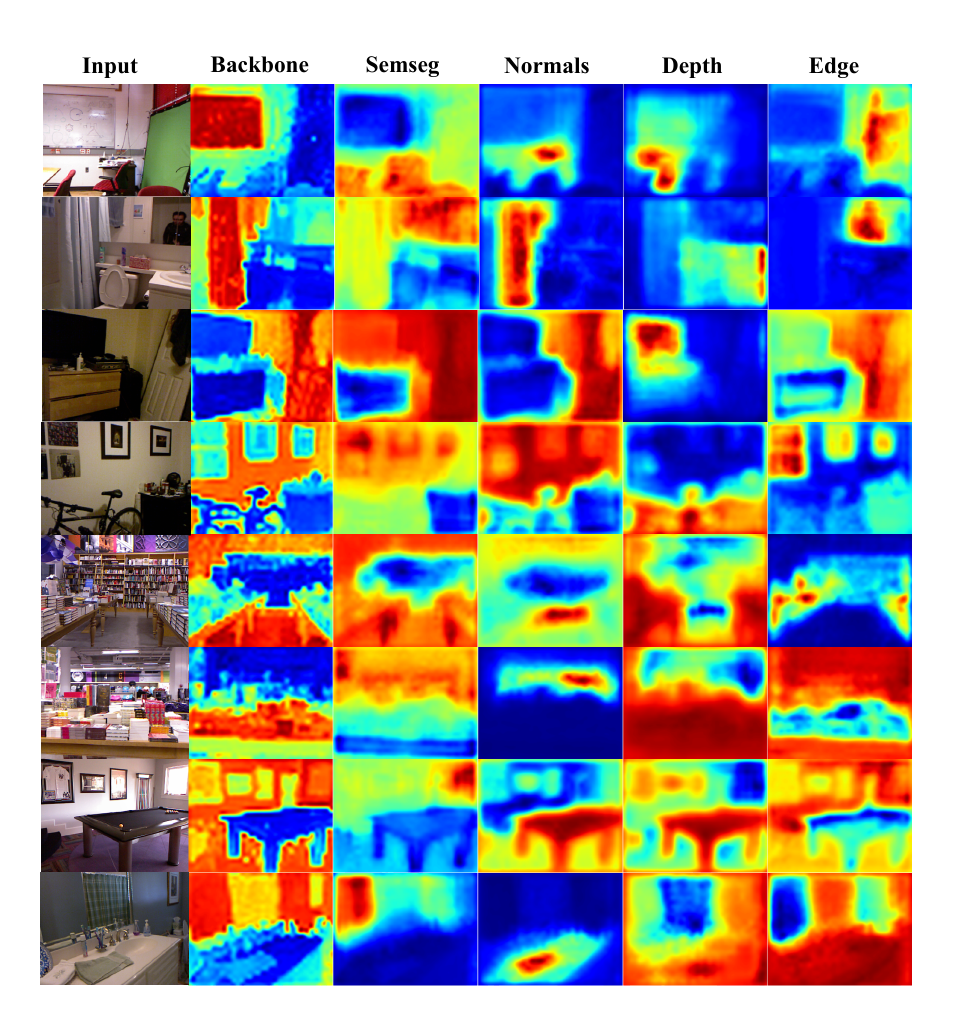}
    \caption{The feature visualization comparison on NYUD-v2. From left to right are the input image, the PCA visualization of the general feature from the final layer of the backbone, and the PCA visualizations of the four task-specific features refined by the PAME module.}
    \label{fig:nyu_self_pca} 
\end{figure}

We visually compared prediction results from our method against\linebreak[4]InvPT++~\cite{ye2024invpt++} on two datasets, as illustrated in  Fig.~\ref{fig:pascal_vis} and Fig.~\ref{fig:nyu_vis}. Our method outperformed InvPT++ in scenarios requiring the interpretation of large-scale semantic contexts and in capturing finer details. It is evident that our baseline model, which employs a single task-specific decoder head, struggled under high task competition, particularly in both fine detail and large semantic regions, compared to our proposed method. Furthermore, we performed PCA dimensionality reduction on the decoding features within our Task Decoder heads, as depicted in Fig.~\ref{fig:pca}. The analysis reveals that the baseline model, lacking comprehensive interaction mechanisms, exhibited considerable high-frequency noise, whereas our method delivered more stable outputs. In extensive semantic areas, our method consistently produced more coherent and reliable feature representations. We compared the features before and after applying the PAME module, as illustrated in Fig.~\ref{fig:pascal_self_pca} and Fig.~\ref{fig:nyu_self_pca}. The results demonstrate that the PAME module's interactive processing enables task-specific adaptations of general features. For example, in human segmentation, the focus shifts to human limbs, whereas in depth estimation, the emphasis is on the distinction between proximal foregrounds and distant backgrounds.

\subsection{Ablation Studies and Analysis}
To validate the efficacy of various components, we conducted extensive ablation studies using the PASCAL-Context dataset with a vision Transformer (ViT) backbone. The baseline configuration employs a ViT-B model featuring 12 layers. For multi-scale feature extraction, we selected side outputs from layers 3, 6, 9, and 12 of the backbone. These features were directly projected to the respective output channels corresponding to each task using a task-specific decoder.
\begin{figure*}
    \centering
    \includegraphics[width=1.0\linewidth]{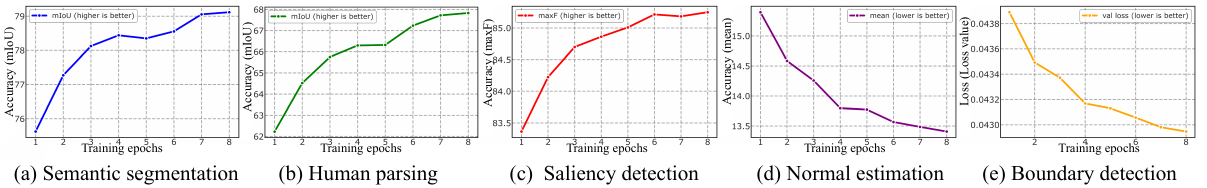}
    \caption{The impact of epoch changes on experimental results. Experiments on PAME using the ViT-L backbone on PASCAL-Context.}
    \label{fig:pascal_epoch}
\end{figure*}
\subsubsection{Ablative Experiments of The Main Components}
\begin{table}[tp]
    \setlength{\tabcolsep}{3pt}
    \caption{Ablation study of parameter prior (PP) on the PASCAL-Context dataset, including two parameters}
    \label{tab:ablation PP}
    \vspace{2mm}
    \centering
    \scalebox{1}{
    \renewcommand{\arraystretch}{1.7}
        \begin{tabular}{
        >{\centering\arraybackslash}p{0.5cm}
        >{\centering\arraybackslash}p{0.5cm}
        >{\centering\arraybackslash}p{1cm}
        >{\centering\arraybackslash}p{1cm}
        >{\centering\arraybackslash}p{1cm}
        >{\centering\arraybackslash}p{1cm}
        >{\centering\arraybackslash}p{1cm}
        >{\centering\arraybackslash}p{1cm}}
        \thickhline
        \multicolumn{2}{c}{\textbf{PP}}   & \multicolumn{1}{|c}{\textbf{Semseg}} & \textbf{Parsing} & \textbf{Saliency} & \textbf{Normal} & \textbf{Boundary} &\textbf{$\Delta_{g}$}\\[-1.8ex]
        
         \textbf{B}& \textbf{C}& \multicolumn{1}{|c}{mIoU$\uparrow$} & mIoU$\uparrow$ & maxF$\uparrow$ & mErr$\downarrow$ & odsF$\uparrow$  &\%$\uparrow$ \\
        \hline
        $\checkmark$&$\checkmark$& \multicolumn{1}{|c}{\textbf{79.12}}& \textbf{67.83}& \textbf{85.25}& \textbf{13.40}&     \textbf{73.40}&\textbf{0.49}\\[-1.5ex]
     
        \ding{55}&$\checkmark$& \multicolumn{1}{|c}{\underline{78.82}}& 67.52& \underline{84.98}& 13.43& \underline{73.20}&\underline{0.16}\\[-1.5ex]
 
        $\checkmark$& \ding{55}& \multicolumn{1}{|c}{78.72}& \underline{67.71}& 84.96& \underline{13.42}& \underline{73.20}&0.20\\[-1.5ex]
 
        \ding{55}& \ding{55}& \multicolumn{1}{|c}{78.80}& 67.46& 84.94& 13.50& 73.10&0.00\\
        \thickhline
  \end{tabular}}
\end{table}
\begin{table}[tp]
    \setlength{\tabcolsep}{3pt}
    \caption{Comparison of the effects of different top-k values on the PASCAL-Context dataset.}
    \label{tab:ablation topk}
    
    \vspace{2mm}
    \centering
    \scalebox{1}{
    \renewcommand{\arraystretch}{1.7}
        \begin{tabular}{
        >{\centering\arraybackslash}p{0.6cm}
        >{\centering\arraybackslash}p{1cm}
        >{\centering\arraybackslash}p{1cm}
        >{\centering\arraybackslash}p{1cm}
        >{\centering\arraybackslash}p{1cm}
        >{\centering\arraybackslash}p{1cm}
        >{\centering\arraybackslash}p{1cm}}
        \thickhline
        \multirow{2}{*}[0.1cm]{\textbf{K}}   & \multicolumn{1}{|c}{\textbf{Semseg}} & \textbf{Parsing} & \textbf{Saliency} & \textbf{Normal} & \textbf{Boundary} &\textbf{$\Delta_{g}$}\\[-1.8ex]

        & \multicolumn{1}{|c}{mIoU$\uparrow$} & mIoU$\uparrow$ & maxF$\uparrow$ & mErr$\downarrow$ & odsF$\uparrow$  &\%$\uparrow$ \\
        
        \hline
        1& \multicolumn{1}{|c}{78.22}& 66.75& 85.27& 13.51& 72.80& -0.87\\[-1.5ex]
        3& \multicolumn{1}{|c}{78.53}& 66.50& 84.99& 13.51& 73.00& -0.88\\[-1.5ex]
        5& \multicolumn{1}{|c}{78.74}& 67.24& 85.09& 13.51& 73.20& -0.53\\[-1.5ex]
        7& \multicolumn{1}{|c}{78.47}& \underline{67.58}& \textbf{85.39}& 13.50& \underline{73.30}& \underline{-0.38}\\[-1.5ex]
        9& \multicolumn{1}{|c}{\textbf{79.12}}& \textbf{67.83}& \underline{85.25}& \underline{13.40}& \textbf{73.40}& \textbf{0.00}\\[-1.5ex]
        11& \multicolumn{1}{|c}{78.20}& 66.84& 85.07& \textbf{13.38}& \textbf{73.40}& -0.54\\[-1.5ex]
        13& \multicolumn{1}{|c}{78.48}& 66.85& 85.15& 13.54& 73.20& -0.74\\[-1.5ex]
        15& \multicolumn{1}{|c}{\underline{79.08}}& 67.60& 85.15& 13.72& 73.1& -0.82\\
        
        \thickhline
    \end{tabular}}
\end{table}

We conducted an ablation study on the two primary components of our proposed method. Results from TABLE~\ref{tab:ablation main block} show that integrating Task Convolution (TC) leads to a modest performance improvement, focusing primarily on local task features. However, the inclusion of the PAME module further enhances performance, highlighting the benefits of global modeling. Specifically, compared to our baseline, TC alone resulted in a 1.03\% increase in $\Delta_{g}$. Furthermore, integrating the PAME module increased $\Delta_{g}$ by 2.76\%, surpassing the gains achieved with TC alone.

\vspace{1cm}
\begin{table}[tp]
  \setlength{\tabcolsep}{3pt}
  \caption{Ablation study of different scanning methods on the PASCAL-Context dataset.}
  \label{tab:ablation scan}
  \vspace{2mm}
  \centering
  \scalebox{1}{
  \renewcommand{\arraystretch}{1.7}
        \begin{tabular}{
        >{\centering\arraybackslash}p{1cm}
        >{\centering\arraybackslash}p{1cm}
        >{\centering\arraybackslash}p{1cm}
        >{\centering\arraybackslash}p{1cm}
        >{\centering\arraybackslash}p{1cm}
        >{\centering\arraybackslash}p{1cm}
        >{\centering\arraybackslash}p{1cm}}
        \thickhline

        \multirow{2}{*}[0.1cm]{\textbf{Method}}     & \multicolumn{1}{|c}{\textbf{Semseg}} & \textbf{Parsing} & \textbf{Saliency} & \textbf{Normal} & \textbf{Boundary} &\textbf{$\Delta_{g}$} \\[-1.8ex]
        & \multicolumn{1}{|c}{mIoU}$\uparrow$ & mIoU$\uparrow$ & maxF$\uparrow$ & mErr$\downarrow$ & odsF$\uparrow$  &\%$\uparrow$ \\
        \hline
         S-scan& \multicolumn{1}{|c}{78.57}& 67.68& \textbf{85.28}& 13.56& \textbf{73.50}&0.00\\[-1.5ex]
        Z-scan& \multicolumn{1}{|c}{\underline{79.01}}& \underline{67.77}& 84.98& \underline{13.45}& \underline{73.40}&\underline{0.20}\\[-1.5ex]
        H-scan& \multicolumn{1}{|c}{\textbf{79.12}}& \textbf{67.83}& \underline{85.25}& \textbf{13.40}& \underline{73.40}&\textbf{0.39}\\
        \thickhline
      \end{tabular}}
\end{table}
\vspace{1cm}
\begin{table}[tp]
    \setlength{\tabcolsep}{3pt}
    \caption{Ablation study of the number of using stages.}
    \label{tab:ablation stage}
    
    \vspace{2mm}
    \centering
    \scalebox{1}{
    \renewcommand{\arraystretch}{1.7}
        \begin{tabular}{
        >{\centering\arraybackslash}p{1cm}
        >{\centering\arraybackslash}p{1cm}
        >{\centering\arraybackslash}p{1cm}
        >{\centering\arraybackslash}p{1cm}
        >{\centering\arraybackslash}p{1cm}
        >{\centering\arraybackslash}p{1cm}
        >{\centering\arraybackslash}p{1cm}}
        \thickhline
        \multirow{2}{*}[0.1cm]{\textbf{Stage(s)}}   & \multicolumn{1}{|c}{\textbf{Semseg}} & \textbf{Parsing} & \textbf{Saliency} & \textbf{Normal} & \textbf{Boundary} &\textbf{$\Delta_{g}$}\\[-1.8ex]

        & \multicolumn{1}{|c}{mIoU$\uparrow$} & mIoU$\uparrow$ & maxF$\uparrow$ & mErr$\downarrow$ & odsF$\uparrow$  &\%$\uparrow$ \\
        
        \hline
        1& \multicolumn{1}{|c}{75.57}& 65.40& 85.10& 13.96&  72.60 & 0.00\\[-1.5ex]
        2& \multicolumn{1}{|c}{76.94}& 65.63& \underline{85.20}& 13.93& 73.00 &0.61\\[-1.5ex]
        3& \multicolumn{1}{|c}{\underline{78.61}}& \underline{67.17}& 85.15& \underline{13.48}& \underline{73.20}&\underline{2.21}\\[-1.5ex]
        4& \multicolumn{1}{|c}{\textbf{79.12}}&\textbf{67.83}& \textbf{85.25}& \textbf{13.40}&\textbf{73.40}&\textbf{2.74}\\
        \thickhline
    \end{tabular}}
\end{table}

\vspace{-18mm}
\subsubsection{Effectiveness of Parameter Experts}
We performed an ablation study to assess the impact of the Parameter Experts (PE) module by comparing the full module with versions where MoE branches were replaced by linear projection layers. As shown in TABLE~\ref{tab:ablation PE}, removing the MoE branches weakened inter-task interactions, leading to reduced performance in multi-task scenarios. Reintroducing the full structures of PE components $\mathbf{B}$ and $\mathbf{C}$ independently improved $\Delta_{g}$ by 0.19\% and 0.32\%, respectively. When both modules were restored, the combined improvement reached 0.52\%.

\subsubsection{Effectiveness of Parameter Priors}
We investigated the role of task-specific priors in enhancing model performance by conducting an ablation study on parameters $\mathbf{B}$ and $\mathbf{C}$ (TABLE~\ref{tab:ablation PP}). Incorporating these priors enhanced task differentiation and overall performance. Specifically, adding multi-task priors for $\mathbf{B}$ and $\mathbf{C}$ individually improved $\Delta_g$ by 0.16\% and 0.20\%, respectively. When both priors were included, the total improvement reached 0.49\%.

\subsubsection{Impact of Top-k Value} 
We examined the effect of varying the top-k value in the Parameter Experts (TABLE~\ref{tab:ablation topk}). The results demonstrate that increasing the k-value significantly enhances performance by better leveraging inter-task information. However, beyond an optimal threshold, performance starts to degrade, indicating potential negative transfer effects in parameters $\mathbf{B}$ and $\mathbf{C}$. This analysis helped determine the optimal k-value for our model.

\subsubsection{Comparison of Different Scanning Mechanisms}
We evaluated four scanning directions and three scanning methods (TABLE~\ref{tab:ablation scan}). For the Z-scan, based on the methodology in~\cite{huang2024localmamba}, we established a primary scanning direction and rotated it to create four distinct paths. The S-scan, implemented following \cite{yang2024plainmamba}, also employed four directions. Experimental results indicate that the H-scan outperformed the other methods, likely due to its superior spatial locality and continuity compared to both the S-scan and Z-scan.

\subsubsection{Influence of Training Epochs}
We analyzed the impact of varying training epochs on performance (Fig.~\ref{fig:pascal_epoch}), recording metrics at each epoch. Initially, the model exhibited underfitting, with consistent improvements during the first four epochs. By the fifth epoch, most tasks reached a plateau or showed minor declines, with the exception of saliency detection, which continued to improve. This suggests peak task competition occurred at this point. From the sixth epoch onward, as saliency detection converged, task competition eased, leading to overall performance improvements. Our method demonstrated a steady upward trend without significant negative transfer across tasks.

\subsubsection{Effect of Selected Stages} We studied the impact of selecting different stages of the ViT model for feature extraction (TABLE~\ref{tab:ablation stage}). We focused on the final stages and varied the number of selected stages. Results show that increasing the number of stages substantially improves performance. Specifically, using all four stages, compared to just one, yielded a 2.74\% increase in $\Delta_g$.

\begin{table}[h!]
    \setlength{\tabcolsep}{3pt}
    \caption{Ablation study of the number of the direction of MDHS on the PASCAL-Context dataset, including two parameters.}
    \label{tab:ablation scan_num}
    
    \vspace{2mm}
    \centering
    \scalebox{0.98}{
        \renewcommand{\arraystretch}{1.7}
        \begin{tabular}{
        >{\centering\arraybackslash}p{1cm}
        >{\centering\arraybackslash}p{1cm}
        >{\centering\arraybackslash}p{1cm}
        >{\centering\arraybackslash}p{1cm}
        >{\centering\arraybackslash}p{1cm}
        >{\centering\arraybackslash}p{1cm}
        >{\centering\arraybackslash}p{1cm}}
        \thickhline
        \multirow{2}{*}{\textbf{Number}} & \multicolumn{1}{|c}{\textbf{Semseg}} & \textbf{Parsing} & \textbf{Saliency} & \textbf{Normal} & \textbf{Boundary} &\textbf{$\Delta_{g}$}\\[-1.8ex]
        & \multicolumn{1}{|c}{mIoU$\uparrow$} & mIoU$\uparrow$ & maxF$\uparrow$ & mErr$\downarrow$ & odsF$\uparrow$  & \%$\uparrow$ \\
        \hline
        $1$ & \multicolumn{1}{|c}{78.08}& 64.90& 85.16& 13.92& \underline{73.20}& -1.98\\[-1.5ex]
        $2$ & \multicolumn{1}{|c}{\underline{78.22}}& \underline{66.69}& \underline{85.20}& \underline{13.78}& \underline{73.20}& -1.20\\[-1.5ex]
        $4$ & \multicolumn{1}{|c}{\textbf{79.12}}& \textbf{67.83}& \textbf{85.25}& \textbf{13.40} &  \textbf{73.40} & \textbf{0.00}\\
        \thickhline
    \end{tabular}}
\end{table}

\subsubsection{Comparison of Different Scanning Numbers} We conducted comparative experiments with different numbers of scans in MDHS. Due to the difference between Mamba's sequential causal modeling characteristics and the spatial properties of 2D images, it is necessary to design multi-directional scanning methods to enhance the model's perceptual ability. As shown in the TABLE~\ref{tab:ablation scan_num}, when we reduced the number of scans in MDHS to 1 and 2, the model's performance significantly deteriorated, demonstrating the effectiveness of the multi-directional scanning method used.

\section{Conclusion}
In this paper, we presented PAMM, a simple yet effective approach for multi-task dense prediction. To address the challenges of global multi-task perception and task-specific prior learning, we introduced key modifications to the standard Mamba block. First, to enhance cross-task information exchange, we integrated Mixture-of-Experts (MoE) blocks into Mamba’s parameter computation, enabling dynamic, task-adaptive parameterization. Second, to better capture task-specific priors, we incorporated task-specific priors into Mamba’s parameters, facilitating the learning of unique task properties. Additionally, we proposed a multi-directional scanning technique to improve Mamba’s ability to capture 2D spatial features by transforming 2D data into optimized 1D sequences. This method outperforms alternative reconstruction strategies. The effectiveness of PAMM was demonstrated through comprehensive quantitative and qualitative evaluations on two benchmark datasets.






\bibliographystyle{IEEEtran}
\bibliography{reference}

\vfill

\end{document}